# Independence of Causal Influence and Clique Tree Propagation

Nevin Lianwen Zhang and Li Yan
Department of Computer Science, Hong Kong University of Science & Technology
{lzhang, yanli}@cs.ust.hk

## Abstract

This paper explores the role of independence of causal influence (ICI) in Bayesian network inference. ICI allows one to factorize a conditional probability table into smaller pieces. We describe a method for exploiting the factorization in clique tree propagation (CTP) — the state-of-the-art exact inference algorithm for Bayesian networks. We also present empirical results showing that the resulting algorithm is significantly more efficient than the combination of CTP and previous techniques for exploiting ICI.

**Keywords:** Bayesian networks, independence of causal influence (causal independence), inference, clique tree propagation.

## 1  INTRODUCTION

Bayesian networks (Pearl [16], Howard and Matheson [8]) are a knowledge representation framework widely used by AI researchers for reasoning under uncertainty. They are directed acyclic graphs where each node represents a random variable and is associated with a conditional probability table of the node given its parents. This paper is about inference in Bayesian networks. There exists a rich collection of algorithms. The state-of-the-art is an exact algorithm called clique tree propagation[1] (CTP) (Lauritzen and Spiegelhalter [12], Jensen et al [10], and Shafer and Shenoy [20]).

Unfortunately, there are applications that CTP cannot deal with or where it is too slow (e.g. [18]). Much recent effort has been spent on speeding up inference. The efforts can be classified into those that approximate (e.g. [15], [2], [9], [6], [7], [17], [22], [11], and [19])
and those that exploit structures in the probability tables (e.g. [3], [1]).

We are interested in exploiting structures in the probability tables induced by independence of causal influence (ICI). The concept of ICI was first introduced by Heckerman [3] under the name causal independence. It refers to the situation where multiple causes independently influence a common effect. We use the term "independence of causal influence" instead of "causal independence" because many researchers have come to agree that it captures the essence of the situation better than the latter.

Knowledge engineers had been using specific models of ICI in simplifying knowledge acquisition even before the inception of the concept ([5], [13]). Olesen et al [13] and Heckerman [3] have also shown how ICI can be used to simplify the structures of Bayesian networks so that inference can be more efficient.

Zhang and Poole ([23]) made the observation that ICI enables one to factorize a conditional probability table into smaller pieces and showed how the VE algorithm — another exact inference algorithm — can be extended to take advantage of the factorization. This paper extends CTP to exploit conditional probability table factorization. We also present empirical results showing that the extended CTP is more efficient than the combination of CTP and the network simplification techniques. In comparison with Zhang and Poole [23], this paper presents a deeper understanding of ICI. The theory is substantially simplified.

## 2  BAYESIAN NETWORKS

A *Bayesian network* (BN) is an annotated directed acyclic graph, where each node represents a random variable and is attached with a conditional probability of the node given its parents. In addition to the explicitly represented conditional probabilities, a BN also implicitly represents conditional independence as-

---
[1] Also known as junction tree propagation.

sertions. Let $x_1, x_2, \ldots, x_n$ be an enumeration of all the nodes in a BN such that each node appears before its children, and let $\pi_{x_i}$ be the set of parents of a node $x_i$. The following assertions are implicitly represented:

For $i=1, 2, \ldots n$, $x_i$ is conditionally independent of variables in $\{x_1, x_2, \ldots, x_{i-1}\} \setminus \pi_{x_i}$ given variables in $\pi_{x_i}$.

The conditional independence assertions and the conditional probabilities together entail a joint probability over all the variables. As a matter of fact, by the chain rule, we have

$$P(x_1, x_2, \ldots, x_n) = \prod_{i=1}^{n} P(x_i|x_1, x_2, \ldots, x_{i-1})$$
$$= \prod_{i=1}^{n} P(x_i|\pi_{x_i}), \quad (1)$$

where the second equation is true because of the conditional independence assertions and the conditional probabilities $P(x_i|\pi_{x_i})$ are given in the specification of the BN. Consequently, one can, in theory, do arbitrary probabilistic reasoning in a BN.

## 3  INDEPENDENCE OF CAUSAL INFLUENCE

Bayesian networks place no restriction on how a node depends on its parents. Unfortunately this means that in the most general case we need to specify an exponential (in the number of parents) number of conditional probabilities for each node. There are many cases where there is structure in the probability tables. One such case that we investigate in this paper is known as independence of causal influence (ICI).

The concept of ICI was first introduced by Heckerman [4]. The following definition first appeared in Zhang and Poole [24].

In one interpretation, arcs in a BN represent causal relationships; the parents $c_1, c_2, \ldots, c_m$ of a node $e$ are viewed as causes that jointly bear on the effect $e$. ICI refers to the situation where the causes $c_1, c_2 \ldots,$ and $c_m$ contribute independently to the effect $e$. In other words, the ways by which the $c_i$'s influence $e$ are independent.

More precisely, $c_1, c_2 \ldots,$ and $c_m$ are said to *influence $e$ independently* if there exist random variables $\xi_1, \xi_2 \ldots,$ and $\xi_m$ that have the same *frame* — set of possible values — as $e$ such that

1. For each $i$, $\xi_i$ probabilistically depends on $c_i$ and is conditionally independent of all other $c_j$'s and all other $\xi_j$'s given $c_i$, and

2. There exists a commutative and associative binary operator $*$ over the frame of $e$ such that $e = \xi_1 * \xi_2 * \ldots * \xi_m$.

We shall refer to $\xi_i$ as the *contribution* of $c_i$ to $e$. In less technical terms, causes influence their common effect independently if individual contributions from different causes are independent and the total influence is a combination of the individual contributions.

We call the variable $e$ a *convergent variable* for it is where independent contributions from different sources are collected and combined (and for the lack of a better name). Non-convergent variables will simply be called *regular variables*. We also call $*$ the *base combination operator* of $e$. Different convergent variables can have difference base combination operators.

The reader is referred to [24] for more detailed explanations and examples of ICI.

The conditional probability table $P(e|c_1, \ldots, c_m)$ of a convergent variable $e$ can be factorized into smaller pieces. To be more specific, let $f_i(e, c_i)$ be the function defined by

$$f_i(e=\alpha, c_i) = P(\xi_i=\alpha|c_i),$$

for each possible value $\alpha$ of $e$. It will be referred to as the *contributing factor* of $c_i$ to $e$. Zhang and Poole [24] have shown that

$$P(e|c_1, \ldots, c_m) = \otimes_{i=1}^{m} f_i(e, c_i),$$

where $\otimes$ is an operator for combining factors to be defined in the following.

Assume there is a fixed list of variables, some of which are designated to be convergent and others are designated to be regular. We shall only consider functions of variables on the list.

Let $f(e_1, \ldots, e_k, A, B)$ and $g(e_1, \ldots, e_k, A, C)$ be two functions that share convergent variables $e_1, \ldots, e_k$ and a list $A$ of regular variables. $B$ is the list of variables that appear only in $f$, and $C$ is the list of variables that appear only in $g$. Both $B$ and $C$ can contain convergent variables as well as regular variables. Suppose $*_i$ is the base combination operator of $e_i$. Then, the *combination $f \otimes g$ of $f$ and $g$* is a function of variables $e_1, \ldots, e_k$ and of the variables in $A$, $B$, and $C$. It is defined by

$$f \otimes g(e_1=\alpha_1, \ldots, e_k=\alpha_k, A, B, C)$$
$$= \sum_{\alpha_{11} *_1 \alpha_{12} = \alpha_1} \cdots \sum_{\alpha_{k1} *_k \alpha_{k2} = \alpha_k}$$
$$f(e_1=\alpha_{11}, \ldots, e_k=\alpha_{k1}, A, B)$$
$$g(e_1=\alpha_{12}, \ldots, e_k=\alpha_{k2}, A, C), \quad (2)$$

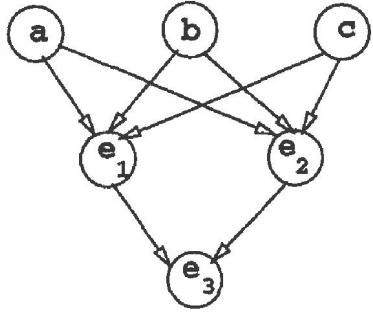

Figure 1: A Bayesian network.

for each possible value $\alpha_i$ of $e_i$. We shall sometimes write $f \otimes g$ as $f(e_1, \ldots, e_k, A, B) \otimes g(e_1, \ldots, e_k, A, C)$ to make explicit the arguments of $f$ and $g$.

The operator $\otimes$ is associative and commutative. When $f$ and $g$ do not share convergent variables, $f \otimes g$ is simply the multiplication $fg$.

## 4 FACTORIZATION OF JOINT PROBABILITIES

A BN represents a factorization of a joint probability. For example, the Bayesian network in Figure 1 factorizes the joint probability $P(a, b, c, e_1, e_2, e_3)$ into the following list of *factors*:

$$P(a), P(b), P(c), P(e_1|a,b,c), P(e_2|a,b,c), P(e_3|e_1,e_2).$$

The joint probability can be obtained by multiplying the factors. We say that this factorization is *multiplication-homogeneous* because all the factors are combined in the same way by multiplication.

Now suppose the $e_i$'s are convergent variables. Then their conditional probabilities can be further factorized as follows:

$$\begin{aligned} P(e_1|a,b,c) &= f_{11}(e_1,a) \otimes f_{12}(e_1,b) \otimes f_{13}(e_1,c), \\ P(e_2|a,b,c) &= f_{21}(e_2,a) \otimes f_{22}(e_2,b) \otimes f_{23}(e_2,c), \\ P(e_3|e_1,e_2) &= f_{31}(e_3,e_1) \otimes f_{32}(e_3,e_2), \end{aligned}$$

where the factor $f_{11}(e_1, a)$, for instance, is the contributing factor of $a$ to $e_1$.

We say that the following list of factors

$$\begin{aligned} &f_{11}(e_1,a), f_{12}(e_1,b), f_{13}(e_1,c), \\ &f_{21}(e_2,a), f_{22}(e_2,b), f_{23}(e_2,c), \\ &f_{31}(e_3,e_1), f_{32}(e_3,e_2), \\ &P(a), P(b), \text{ and } P(c) \end{aligned}$$

constitute a *heterogeneous factorization* of $P(a, b, c, e_1, e_2, e_3)$ because the joint probability can be obtained by combining those factors in a proper order using either multiplication or the operator $\otimes$. The word heterogeneous is to signify the fact that different factor pairs might be combined in different ways. We shall refer to the factorization as the heterogeneous factorization represented by the BN in Figure 1.

The heterogeneous factorization is of finer grain than the homogeneous factorization. The purpose of this paper is to exploit such finer-grain factorizations to speed up inference.

## 5 DEPUTATION

In a heterogeneous factorization, the order by which factors can be combined is rather restrictive. The contributing factors of a convergent variable must be combined with themselves before they can be multiplied with other factors. This is the main issue that we need to deal with in order to take advantage of conditional probability table factorizations induced by ICI.

To alleviate the problem, we introduce the concept of deputation. It was originally defined in term of BNs [24]. In this paper, we define it in terms of heterogeneous factorizations themselves.

In the heterogeneous factorization represented by a BN, to *depute* a convergent variable $e$ is to make a copy $e'$ of $e$ and replace $e$ with $e'$ in all the contributing factors of $e$. The variable $e'$ is called the *deputy* of $e$ and it is designated to be convergent. After deputation, the original convergent variable $e$ is no longer convergent and is called a *new regular variable*. In contrast, variables that are regular before deputation are called *old regular* variables.

After deputing all convergent variables, the heterogeneous factorization represented by the BN in Figure 1 becomes the following list of factors:

$$\begin{aligned} &f_{11}(e'_1,a), f_{12}(e'_1,b), f_{13}(e'_1,c), f_{21}(e'_2,a), \\ &f_{22}(e'_2,b), f_{23}(e'_2,c), f_{31}(e'_3,e_1), f_{32}(e'_3,e_2), \\ &P(a), P(b), P(c). \end{aligned}$$

The rest of this section is to show that deputation renders it possible to combine the factors in arbitrary order.

### 5.1 ELIMINATING DEPUTY VARIABLES IN FACTORS

*Eliminating* a deputy variable $e'$ in a factor $f$ means to replace it with the corresponding new regular variable

$e$. The resulting factor will be denoted by $f|_{e'=e}$. To be more specific, for any factor $f(e,e',A)$ of $e$, $e'$ and a list $A$ of other variables,

$$f|_{e'=e}(e=\alpha, A) = f(e=\alpha, e'=\alpha, A),$$

for each possible value $\alpha$ of $e$. For any factor $f(e',A)$ of $e'$ and a list $A$ of other variables not containing $e$,

$$f|_{e'=e}(e=\alpha, A) = f(e'=\alpha, A),$$

for each possible value $\alpha$ of $e$. For any factor $f$ not involving $e'$, $f|_{e'=e} = f$.

Suppose $f$ involve two deputy variables $e'_1$ and $e'_2$ and we want to eliminate both of them. It is evident that the order by which the deputy variables are eliminated does not affect the resulting factor. We shall denote the resulting factor by $f|_{e'_1=e_1, e'_2=e_2}$.

## 5.2 ⊗-HOMOGENEOUS FACTORIZATIONS

For later convenience, we introduce the concept of ⊗-homogeneous factorization in term of joint potentials. Let $x_1, x_2, \ldots, x_n$ be a list of variables. A *joint potential* $P(x_1, x_2, \ldots, x_n)$ is simply a non-negative function of the variables. Joint probabilities are special joint potentials that sum to one.

Consider a joint potential $P(e_1, \ldots, e_k, x_{k+1}, \ldots, x_n)$ of new regular variables $e_i$ and old regular variables $x_i$. A list of factors $f_1, \ldots, f_m$ of the $e_i$'s, their deputies $e'_i$, and the $x_i$'s is a ⊗-*homogeneous factorization* (reads circle cross homogeneous factorization) of $P(e_1, \ldots, e_k, x_{k+1}, \ldots, x_n)$ if

$$P(e_1, \ldots, e_k, x_{k+1}, \ldots, x_n) = (\otimes_{i=1}^m f_i)|_{e'_1=e_1, \ldots, e'_k=e_k}.$$

**Theorem 1** *Let $\mathcal{F}$ be the heterogeneous factorization represented by a BN and let $\mathcal{F}'$ be the list of factors obtained from $\mathcal{F}$ by deputing all convergent variables. Then $\mathcal{F}'$ is a ⊗-homogeneous factorization of the joint probability entailed by the BN.*

All proofs are omitted due to space limit. Since the operator ⊗ is commetative and associative, the theorem states that factors can be combined in arbitrary order after deputation.

## 6  SUMMING OUT VARIABLES

Summing out a variable from a factorization is a fundamental operation in many inference algorithms. This section shows how to sum out a variable from a ⊗-homogeneous factorization of a joint potential.

Let $\mathcal{F}$ be a ⊗-homogeneous factorization of a joint potential $P(x_1, x_2, \ldots, x_n)$. Consider the following procedure.

Procedure sumoutc($\mathcal{F}, x_1$)

1. **If** $x_1$ is a new regular variable, remove from $\mathcal{F}$ all the factors that involve the deputy $x'_1$ of $x_1$, combine them by ⊗ resulting in, say, $f$. Add the new factor $f|_{x'_1=x_1}$ to $\mathcal{F}$. **Endif**
2. Remove from $\mathcal{F}$ all the factors that involve $x_1$, combine them by using ⊗ resulting in, say, $g$. Add the new factor $g$ to $\mathcal{F}$.
3. Return $\mathcal{F}$.

**Theorem 2** *The list of factors returned by sumoutc($\mathcal{F}, x_1$) is a ⊗-homoogeneous factorization of $P(x_2, \ldots, x_n) = \sum_{x_1} P(x_1, x_2, \ldots, x_n)$.*

## 7  MODIFYING CLIQUE TREE PROPAGATION

Theorem 2 allows one to exploit ICI in many inference algorithms, including VE and CTP. This paper shows how CTP can be modified to take advantage of the theorem. The modified algorithm will be referred to as CTPI. As CTP, CTPI consists of five steps; namely clique tree construction, clique tree initialization, evidence absorption, propagation, and posterior probability calculation. We shall discuss the steps one by one. Familiarity with CTP is assumed.

### 7.1  CLIQUE TREE CONSTRUCTION

A *clique* is simply a subset of nodes. A *clique tree* is a tree of cliques such that if a node appear in two different cliques then it appears in all cliques on the path between those two cliques.

A clique tree for a BN is constructed in two steps: first obtain an undirected graph and then build a clique tree for the undirected graph. CTPI and CTP differ only in the first step. CTP obtains an undirected graph by marrying the parents of each node (i.e. by adding edges between the parents so that they are pairwise connected) and then drop directions on all arcs. The resulting undirected graph is called a moral graph of the BN.

In CTPI, only the parents of regular nodes (representing old regular variables) are married. The parents of convergent nodes (representing new regular variables) are not married. The clique tree constructed in CTPI has the following properties: (1) for any regular node there is a clique that contain the node as well as all its parents and (2) for any convergent node $e$ and each of its parents $x$ there is a clique that contains both $e$ and $x$.

## 7.2 CLIQUE TREE INITIALIZATION

CTPI initializes a clique tree as follows:

1. For each regular node, find *one* clique that contains the node as well as all its parents and attach the conditional probability of the node to that clique.

2. For each convergent node $e$

   (a) If there is a clique that contains the node and all its parents, regard $e$ as a regular node and proceed as in step 1.

   (b) Otherwise for each parent $x$ of $e$, let $f(x,e)$ be the contributing factor of $x$ to $e$. Find *one* clique that contains both $e$ and $x$, attached to that clique the factor $f(x,e')$, where $e'$ is the deputy of $e$.

After initialization, a clique is associated with a list (possibly empty) of factors.

A couple of notes are in order. Factorizing the conditional probability table of a convergent variable $e$ into smaller pieces can bring about gains in inference efficiency because the smaller pieces can be combined with other factors before being combined with themselves, resulting in smaller intermediate factors. If there is a clique that contains $e$ and all its parents, then all the smaller pieces are combined at the same time when processing the clique. In such a case, we are better off to regard $e$ as a regular node (representing an old regular variable).

Second, in CTP all factors associated with a clique are combined at initialization and the resulting factor still involves only those variables in the clique. It is not advisable to do the same in CTPI because the factors involve not only variables in the clique but also deputies of new regular variables in the clique. Combining them all right away can create an unnecessarily large factor and leads to inefficiency. Experiments have confirmed this intuition.

On the other hand, if all variables that appear in one factor $f$ in the list also appear in another factor $g$ in the list, it does not increase complexity to combine $f$ and $g$. Thus we can *reduce* the list by carrying out such combinations. Thereafter, we keep the reduced list of factors and combine a factor with others only when we have to.

Since $\otimes$ is commutative and associative, the factors associated the cliques constitute a $\otimes$-homogenous factorization of the joint probability entailed by the BN.

## 7.3 EVIDENCE ABSORPTION

Suppose a variable $x$ is observed to take value $\alpha$. Let $\chi_{x=\alpha}(x)$ be the function that takes value 1 when $x=\alpha$ and 0 otherwise. CTPI absorbs the piece of evidence that $x=\alpha$ as follows: find all factors that involve $x$ and multiply $\chi_{x=\alpha}(x)$ to those factors.

Let $x_{m+1}, \ldots, x_n$ be all the observed variables and $\alpha_{m+1}, \ldots, \alpha_n$ be their observed values. Let $x_1, \ldots, x_m$ be all unobserved variables. After evidence absorption, the factors associated with the cliques constitute a $\otimes$-homogenous factorization of joint potential $P(x_1, \ldots, x_m, x_{m+1}=\alpha_{m+1}, \ldots, x_n=\alpha_n)$ of $x_1, \ldots, x_m$.

## 7.4 CLIQUE TREE PROPAGATION

Just as in CTP, propagation in CTPI is done in two sweeps. In the first sweep messages are passed from the leaf cliques toward a pivot clique and in the second sweep messages are passed from the pivot clique toward the leaf cliques. Unlike in CTP where messages passed between neighboring cliques are factors, in CTPI messages passed between neighboring cliques are lists of factors.

Let $C$ and $C'$ be two neighboring cliques. Messages can be passed from $C$ to $C'$ when $C$ has received messages from all the other neighbors. Suppose $x_1, \ldots, x_l$ are all the variables in $C \setminus C'$. Let $\mathcal{F}$ be the list of the factors associated with $C$ and the factors sent to $C$ from all other neighbors of $C$. Messages are passed from $C$ to $C'$ by using the following subroutine.

   Procedure sendMessage($C, C'$)

   1. For $i=1$ to $l$, $\mathcal{F} = $ sumoutc($\mathcal{F}, x_i$), Endfor
   2. Reduce the list $\mathcal{F}$ of factors and send the reduced list to $C'$.

## 7.5 POSTERIOR PROBABILITIES

**Theorem 3** *Let $C$ be a clique and let $x_1, \ldots, x_l$ be all unobserved variables in $C$. Then the factors associated with $C$ and the factors sent to $C$ from all its neighbors constitute a $\otimes$-homogeneous factorization of $P(x_1, \ldots, x_l, x_{m+1}=\alpha_{m+1}, \ldots, x_n=\alpha_n)$.*

Because of Theorem 3, the posterior probability of any unobserved variable $x$ can be obtained as follows:

   Procedure getProb($x$)

   1. Find a clique $C$ that contains $x$. (Let $x_2, \ldots, x_l$ be all other variables in $C$. Let $\mathcal{F}$ be the list of the factors associated with

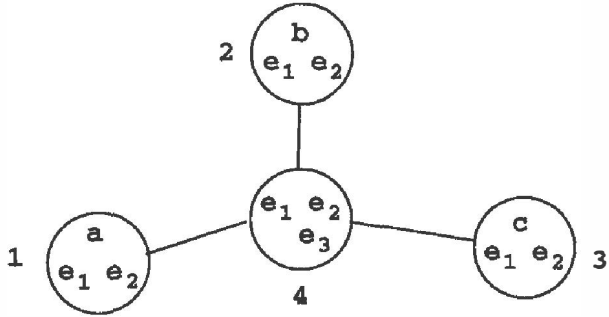

Figure 2: A clique tree for the BN in Figure 1.

$C$ and the factors sent to $C$ from all its neighbors.)

2. **For** $i=2$ to $l$, $\mathcal{F} = \mathtt{sumoutc}(\mathcal{F}, x_i)$, **End-for**

3. Combine all factors in $\mathcal{F}$. Let $f$ be the resulting factor.

4. **If** $x$ is a new regular variable return $f|_{x'=x}/\sum_x f|_{x'=x}$.

5. **Else** return $f/\sum_x f$.

## 8 AN EXAMPLE

A clique tree for the BN in Figure 1 is shown in Figure 2. After initialization, the lists of factors associated with the cliques are as follows:

$$l_1 = \{P(a)f_{11}(e_1', a), f_{21}(e_2', a)\},$$
$$l_2 = \{P(b)f_{12}(e_1', b), f_{22}(e_2', b)\},$$
$$l_3 = \{P(c)f_{13}(e_1', c), f_{23}(e_2', c)\},$$
$$l_4 = \{P(e_3|e_1, e_2)\}.$$

Several factors are combined due to factor list reduction and combination of factors reduces to multiplication because they do not share convergent variables. Also because $e_3$ and all its parents appear in clique 4, its conditional probability is not factorized. It is hence regarded as an old regular variable.

Suppose $e_1$ is observed to take value $\alpha$. Since $P(e_3|e_1, e_2)$ is the only factor that involves $e_1$, absorbing the piece of evidence changes the list of factors associated with clique 4 to the following:

$$l_4 = \{P(e_3|e_1, e_2)\chi_{e_1=\alpha}(e_1)\}.$$

Suppose clique 4 is chosen to be the pivot. Then messages are first propagated from cliques 1, 2, and 3 to clique 4 and then from clique 4 to cliques 1, 2, and 3. The message from clique 1 to clique 4 is obtained by summing out variable $a$ from the list $l_1$ of factors. It is the following list of one factor:

$$\{\mu_{1\to 4}(e_1', e_2')\},$$

where $\mu_{1\to 4}(e_1', e_2') = \sum_a P(a)f_{11}(e_1', a)f_{21}(e_2', a)$. Messages from cliques 2 and 3 to 4 are similar.

To figure out the message from clique 4 to clique 1, we notice that the list of factors associated with clique 4 and sent to clique 4 from cliques 2 and 3 is:

$$\{P(e_3|e_1, e_2)\chi_{e_1=\alpha}(e_1), \mu_{2\to 4}(e_1', e_2'), \mu_{3\to 4}(e_1', e_2')\}.$$

The message is obtained by summing out the variable $e_3$ from the list of factors. Summing out $e_3$ results in a new factor

$$\psi(e_1, e_2) = \sum_{e_3} P(e_3|e_1, e_2)\chi_{e_1=\alpha}(e_1).$$

Hence the message is the following list of factors:

$$\{\mu_{2\to 4}(e_1', e_2')\otimes\mu_{3\to 4}(e_1', e_2'), \psi(e_1, e_2)\}.$$

where the first two factors are combined due to factor list reduction. Messages from clique 4 to cliques 2 and 3 are similar.

Consider computing the posterior probability of $e_3$. The only clique where we can do this computation is clique 4. The list of factors associated with clique 4 and factors sent to clique 4 from all its neighbors is

$$\{P(e_3|e_1, e_2)\chi_{e_1=\alpha}(e_1), \quad \mu_{1\to 4}(e_1', e_2'),$$
$$\mu_{2\to 4}(e_1', e_2'), \mu_{3\to 4}(e_1', e_2')\}$$

There are two variables to sum out, namely $e_1$ and $e_2$. Assume $e_1$ is summed out before $e_2$. The first step in summing out $e'$ is to eliminate $e_1'$, yielding a new factor

$$\phi_1(e_1, e_2') =$$
$$[mu_{1\to 4}(e_1', e_2')\otimes mu_{2\to 4}(e_1', e_2')\otimes mu_{3\to 4}(e_1', e_2')]|_{e_1'=e_1}.$$

Then $e_1$ itself is summed out, yielding a new factor

$$\phi_2(e_2, e_2', e_3) = \sum_{e_1} P(e_3|e_1, e_2)\chi_{e_1=\alpha}(e_1)\phi_1(e_1, e_2').$$

And then $e_2'$ is eliminated, yielding a new factor

$$\phi_3(e_2, e_3) = \phi_2(e_2, e_2', e_3)|_{e_2'=e_2}.$$

And then $e_2$ is summed out, yielding a new factor

$$\phi_4(e_3) = \sum_{e_2} \phi_3(e_2, e_3).$$

Finally,

$$P(e_3|e_1=\alpha) = \frac{\phi_4(e_3)}{\sum_{e_3} \phi_4(e_3)}.$$

## 9  EMPIRICAL COMPARISONS WITH OTHER METHODS

This section empirically compares CTPI with CTP. We also compare CTPI with PD&CTP, the combination of the parent-divorcing transformation [14] and CTP, and with TT&CTP, the combination temporal transformation [4] and CTP.

The CPCS networks [19] are used in the comparisons. They are a good testbed for algorithms that exploits ICI since all non-root nodes are convergent. The networks vary in the number of nodes, and the average number of parents of a node, and the average number of possible values of a node (variable). Their specifications are given in the following table.

| Networks  | NN  | AN-PN | AN-PVN |
|-----------|-----|-------|--------|
| Network 1 | 145 | 1.14  | 2.0    |
| Network 2 | 145 | 1.14  | 2.27   |
| Network 3 | 245 | 1.45  | 2.0    |
| Network 4 | 245 | 1.45  | 2.25   |

NN:       number of nodes;
AN-PN:    average number of parents;
AN-PVN:   average number of possible values of a node.

Since clique tree construction and initialization need to be carried out only once for each network, we shall not compare in detail the complexities of algorithms in those two steps, except saying that they do not differ significantly. Computing posterior probabilities after propagation requires very little resources compared to propagation. We shall concentrate on propagation time.

In standard CTP, incoming messages of a clique are combined in the propagation module after message passing. In CTPI, on the other hand, incoming messages are not combined in the propagation module. For fairness of comparison, the version of CTP we implemented postpones the combination of incoming messages to the module for computing posterior probabilities.

Let us define a *case* to consist of a list of observed variables and their observed values. Propagation time and memory consumption varies from case to case. In the first three networks, the algorithms were tested using 150 randomly generated cases consisting of 5, 10, or 15 observed variables. In the fourth network, only 15 cases were used due to time constraints. Propagation times and maximum memory consumptions across the cases were averaged. The statistics are in Figure 3, where the Y-axises are in logscale. All data were collected using a SPARC20.

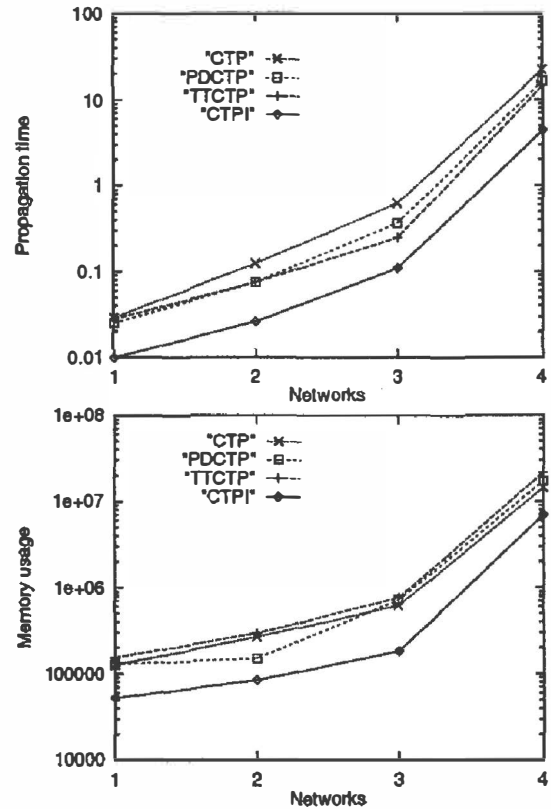

Figure 3: Average space and time complexities of CTP, PD&CTP, TT&CTP, and CTPI on the CPCS networks.

We see that CTPI is faster than all other algorithms and it uses much less memory. In network 4, for instance, CTPI is about 5 faster than CTP, 3 times faster than TT&CTP, and 3.5 times faster than PD&CTP. On average it requires 7MB memory, while CTP requires 15MB, TT&CTP requires 22MB, and PD&CTP require 17MB.

The networks used in our experiments are quite simple in the sense that the nodes have a average number of less than 1.5 parents. As a consequence, gains due to exploitation of ICI and the differences among the different ways of exploiting ICI are not very significant. Zhang and Poole [24] have reported experiments on more complex versions of the CPCS networks with combinations of the VE algorithm and methods for exploiting ICI. Gains due to exploitation of ICI and the differences among the different ways of exploiting ICI are much larger. Unfortunately, none of the combinations of CTP and methods for exploiting ICI was able to deal with those more complex network; they all ran out memory when initializing clique trees.

The method of exploiting ICI described in this paper is more efficient than previous method because it di-

488  Zhang and Yan

rectly takes advantage of the fact that ICI implies conditional probability factorization, while previous methods make use of implications of the fact.

## 10 CONCLUSIONS

We have proposed to method for exploiting ICI in CTP. The method has been empirically shown to be more efficient than the combination of CTP and the network simplification methods for exploiting ICI. Theoretical underpinnings for the method have their roots in Zhang and Poole [24] and are significantly simplified due a deeper understanding of ICI.

### ACKNOWLEDGEMENT

This paper has benefited from discussions with David Poole. Research was supported by Hong Kong Research Council under grant HKUST658/95E and Sino Software Research Center under grant SSRC95/96.EG01.